\documentclass[nonblindrev]{informs3} 

\usepackage{balance}
\usepackage{amsmath}
\usepackage[tight,footnotesize]{subfigure}
\usepackage{graphicx}
\usepackage{bm}
\usepackage{algorithm}
\usepackage[noend]{algorithmic}
\usepackage{multirow}
\usepackage{slashbox}
\usepackage{caption}
\usepackage{amsfonts}
\usepackage{siunitx}

\OneAndAHalfSpacedXII 

\usepackage{natbib}
 \bibpunct[, ]{(}{)}{,}{a}{}{,}%
 \def\bibfont{\small}%
\TheoremsNumberedThrough     

\EquationsNumberedThrough    

\MANUSCRIPTNO{2015} 

\begin{document}


\RUNAUTHOR{Yuan and Tang}

\RUNTITLE{Learning Submodular Sequencing from Samples}

\TITLE{Learning Submodular Sequencing from Samples}

\ARTICLEAUTHORS{%
\AUTHOR{Jing Yuan}
\AFF{Department of Computer Science and Engineering, The University of North Texas}
\AUTHOR{Shaojie Tang}
\AFF{Department of Management Science and Systems, University at Buffalo}
} 

\ABSTRACT{
This paper addresses the problem of sequential submodular maximization: selecting and ranking items in a sequence to optimize some composite submodular function. In contrast to most of the previous works, which assume access to the utility function, we assume that we are given only a set of samples. Each sample includes a random sequence of items and its associated utility. We present an algorithm that, given polynomially many samples drawn from a two-stage uniform distribution, achieves an approximation ratio dependent on the curvature of individual submodular functions. Our results apply in a wide variety of real-world scenarios, such as ranking products in online retail platforms, where complete knowledge of the utility function is often impossible to obtain. Our algorithm gives an empirically useful solution in such contexts, thus proving that limited data can be of great use in sequencing tasks. From a technical perspective, our results extend prior work on ``optimization from samples'' by generalizing from optimizing a set function to a sequence-dependent function.}


\maketitle

\section{Introduction}
Submodular optimization is one of the most important problems in machine learning, with applications in sparse reconstruction \citep{das2011submodular}, data summarization \citep{lin2011class}, active learning \citep{golovin2011adaptive,tang2021optimal}, and viral marketing \citep{tang2020influence}. Most of the existing work is on the problem of selecting a subset of items that maximizes some submodular function. Many real applications, however, require not only the selection of items but also their ranking in a certain order \citep{azar2011ranking,tschiatschek2017selecting,tang2021cascade}.

This paper focuses on one such problem, termed \emph{sequential submodular maximization} \citep{asadpour2022sequential,zhang2022ranking,tang2024non}. The problem's input consists of a ground set $\Omega$ and $k$ submodular functions, denoted as $f_1, \cdots ,f_k: 2^\Omega \rightarrow \mathbb{R}^+$. Our objective is to select a sequence of $k$ items, denoted as $\pi=\{\pi_1, \cdots, \pi_k\}$, from $\Omega$, aiming to maximize
$F(\pi) \stackrel{\text{def}}{=} \sum_{j\in[k]} f_t(\pi_{[t]})$. Here, $\pi_{[t]} \stackrel{\text{def}}{=} \{\pi_1, \cdots, \pi_t\}$ represents the first $t$ items of $\pi$. Notably, each function $f_t$ takes the first $t$ items from the ranking sequence $\pi$ as its input.

This problem captures the position bias in item selection, finding applications in sequential active learning and recommendation systems \citep{zhang2022ranking}. One illustrative example would be product ranking in any of the online retail platforms, like Amazon \citep{asadpour2022sequential}.
Consider Amazon's daily task of selecting and sequencing a list of products, possibly in vertical order, for display to its customers. Customers browse through this list, reaching a certain position, and may proceed to make purchases from the products they view. Then one of the primary objectives of most platforms is to optimize selection and ranking of products to maximize the chance of a purchase. It turns out that this application can be framed as a sequential submodular maximization problem.  In this context, parameters of $F(\pi)$ can be interpreted as follows: Let $ \Omega $ be the set of products and let $k$ be the window size of displayed products. Given a sequence of products $\pi$ of length $k$, for each $ t\in \{1,2, \cdots, k \} $, $f_t(\pi_{[t]})$ is the probability of purchase by customers with patience level $t$, where a customer with a patience level of $t$ would consider viewing the first $t$ products, $\pi_{[t]}$.  Typically, $f_t$ is modeled as a submodular function. In this case, $F(\pi)$ captures the expected purchase probability given that a customer is shown the sequence of products $\pi$.

While sequential submodular maximization has been extensively explored in the literature \citep{asadpour2022sequential}, existing studies typically assume complete knowledge of $f_1, \cdots ,f_k: 2^\Omega \rightarrow \mathbb{R}^+$ and consequently  $F$. However, this assumption is often unrealistic. For instance, in the aforementioned context of recommendation systems, accurately estimating the purchase probability for every product set is often extremely challenging, if not impossible. Instead, a more realistic scenario involves the platform gathering a potentially extensive dataset comprising browsing histories. Each record (a.k.a. sample) within this dataset includes the sequence of displayed products along with customer feedback. For instance, a record could look like this: \{\emph{Sequence}: Product A, Product B; \emph{Feedback}: B was purchased\}. Consequently, the platform aims to identify the best sequence of products based on the samples drawn from some distribution. This problem is highly non-trivial since the platform does not have direct access to the original utility function $F$, making the existing result on submodular sequencing inapplicable.  It has been demonstrated that optimizing a set function from samples is generally impossible, even if the set function is a coverage function \cite{balkanski2017limitations}. Our challenge is compounded by the fact that our function $F$ is defined over a sequence, rather than a set, of items.

Fortunately, in practice, we often encounter submodular functions that may demonstrate more favorable behavior. To this end, we introduce a notation called \emph{curvature} \cite{balkanski2016power}. Intuitively, curvature measures the deviation of a given function from a modular function. Specifically, we say a submodular function $f$ has curvature $c \in [0,1]$ if $f(i\mid S) \geq (1-c)f(\{i\})$ for any $S \subseteq \Omega$ and $i\notin S$. Here  $f(i \mid S)  \stackrel{\text{def}}{=} f(S\cup\{i\})-f(S)$ denotes the marginal utility of an item $i\in \Omega$ on top of a set of items $S\subseteq \Omega$. Hence, if $f$ is a modular function, it has a curvature of $0$.
In general, the complexity of optimizing a submodular function often hinges on the curvature of the focal function. That is, the instances of submodular optimization become challenging typically only when the curvature is unbounded, i.e., $c$ close to $1$. In this paper, we study how to optimize a function $F(\pi) = \sum_{t\in[k]} f_t(\pi_{[t]})$ from samples when the curvature of each individual function $f_t$ is bounded. Our contribution is the development of an approximation algorithm that draws polynomially-many samples from a natural two-stage uniform distribution over feasible sequences and achieves an approximation ratio dependent on the curvature.

\section{Related Work}
While submodular maximization has been extensively studied in the literature \cite{nemhauser1978analysis}, most existing studies assume that the submodular function to be optimized is known. Recently, there has been a line of research focused on learning a submodular function from samples \cite{balcan2011learning,feldman2014learning,balcan2012learning,feldman2013representation}, aiming to construct a function that approximates those from which the samples were collected. It has been shown that monotone submodular functions can be approximately learned from samples drawn from a specific distribution \cite{balcan2011learning}. However, it has also been demonstrated that even if an objective function is learnable from samples, optimization for such a function might still be impossible \cite{balkanski2017limitations}. Despite these negative results, there exists a series of studies \cite{balkanski2016power,chen2020optimization,chen2021network} that develop effective algorithms to optimize submodular functions from samples.

 In our paper, we focus on an important variant of submodular optimization known as  sequential submodular maximization. The objective of sequential submodular maximization is more general than simply selecting a subset of items: it involves jointly selecting and sequencing items. \cite{asadpour2022sequential} studied this problem with monotone and submodular functions. \cite{tang2024non} extended this study to the non-monotone setting. However, all these studies assume a known utility function. Our research builds on and extends these studies by expanding the ``learning-from-samples'' framework \cite{balkanski2016power} from set functions to sequence functions.  Moreover, we identify a gap in the analysis presented in existing studies; more details are provided in  \textbf{Discussions} section.

\section{Preliminaries and Problem Formulation}
\label{sec:pre}
Throughout the remainder of this paper, let  $[m]=\{0, 1, 2, \ldots, m\}$ for any positive integer $m$. Given a function  $f$, let $f(i \mid S)  \stackrel{\text{def}}{=} f(S\cup\{i\})-f(S)$ denote the marginal utility of an item $i\in \Omega$ on top of a set of items $S\subseteq \Omega$.
We say a function $f$ is submodular if and only if for any two sets $X$ and $Y$ such that $X\subseteq Y$ and any item $i\notin Y$, $f(i\mid X) \geq f(i\mid Y)$. Moreover, we say a submodular function $f$ has curvature $c \in [0,1]$ if $f(i\mid S) \geq (1-c)f(\{i\})$ for any $S \subseteq \Omega$ and $i\notin S$.

\subsection{Utility Function} Now we are ready to introduce our research problem. Given $k$ submodular functions $f_1, \cdots ,f_k: 2^\Omega \rightarrow \mathbb{R}^+$, the sequential submodular maximization problem aims to find a sequence $\pi=\{\pi_1, \cdots, \pi_k\}$ from a ground set $\Omega$ that maximizes the value of $F(\pi)$. Here,
\begin{eqnarray}
F(\pi) \stackrel{\text{def}}{=} \sum_{t\in[k]}f_t(\pi_{[t]}),
 \end{eqnarray}
 where $\pi_{[t]} \stackrel{\text{def}}{=} \{\pi_1, \cdots, \pi_t\}$ represents the first $t$ items of $\pi$. That is, each function $f_t$ takes the first $t$ items from $\pi$ as its input. Throughout this paper, we use the notation $\pi$ to denote both a sequence of items and the set of items in that sequence.

Existing studies on sequential submodular maximization all assume that $f_1, \cdots ,f_k$ are known in advance, however, in our setting, we do not have direct access to those functions. Instead, we rely on a dataset comprising observations ${(\pi, \phi(\pi))}$, where in each sample $(\pi, \phi(\pi))$, $\pi$ denotes a feasible sequence and $\phi(\pi)$ denotes the observed utility of $\pi$. It is important to note that the observed utility of a sequence $\pi$ may be subject to randomness, rendering $\phi(\pi)$ a realization of this stochastic variable. Take, for instance, the product sequencing example outlined in the introduction: $F(\pi)$ denotes the likelihood of purchase from a product sequence $\pi$. Here, the observed utility $\phi(\pi)$ of $\pi$ is a binary variable, with $\phi(\pi)=1$ denoting a purchase and $\phi(\pi)=0$ denoting a non-purchase. In this example, randomness stems from two sources: the user's type, characterized by their patience level (i.e., a random function $f_t$ is sampled from $\{f_1, \cdots, f_k\}$), and the probabilistic decision-making process of whether the user will purchase a product from $\pi$ (note that $f_t$ represents only the \emph{aggregated} likelihood of purchase).

\subsection{Problem Formulation} Our objective is to compute a sequence $\pi=\{\pi_1, \cdots, \pi_k\}$ that maximizes the value of $F(\pi)$ based on the samples drawn from a distribution $\mathcal{D}$. We say this problem is $\gamma$-optimizable with respect to a distribution $\mathcal{D}$, if there exists an algorithm which, given polynomially many samples drawn from $\mathcal{D}$, returns with high probability a sequence $\pi$ of size at most $k$ such that
$
F(\pi)\geq \gamma F(\pi^*)
$
where $\pi^*$ denotes the optimal solution of this problem.

 As with the standard \texttt{PMAC}-learning framework, we fix a distribution called \emph{two-stage uniform sampling} and assume that samples are drawn i.i.d. from this distribution. In particular, two-stage uniform sampling works in two stages: In the first stage, a length $t$ is randomly selected from the set $\{1, \cdots, k\}$ with uniform probability. Subsequently, a sequence of length $t$ is randomly chosen, and its realized utility is observed. In the following, we present an approximation algorithm with respect to this distribution.

\section{Algorithm Design}
Our algorithm first estimates the expected marginal contribution $\Delta(i, t)$ of each item $i\in \Omega$ to a uniformly random sequence of size $t$, that does not contain $i$, for every item $i\in \Omega$ and every size $t\in [k-1]$. A formal definition of $\Delta(i, t)$ is given by:
\begin{eqnarray}
\Delta(i, t)= \mathbb{E}_{\Pi_{t+1, i}}\big[F(\Pi_{t+1, i})\big] - \mathbb{E}_{\Pi_{t,-i}}\big[F(\Pi_{t,-i})\big]
\end{eqnarray}
where $\Pi_{t+1, i}$ denotes a random sequence of length $t+1$ with $i$ being placed at the last slot and $\Pi_{t,-i}$ denotes a random sequence of length $t$ that does not contain $i$. Unfortunately, one can not access the value of either $\mathbb{E}_{\Pi_{t+1, i}}\big[F(\Pi_{t+1, i})\big]$ or $\mathbb{E}_{\Pi_{t,-i}}\big[F(\Pi_{t,-i})\big]$ directly.  To estimate these values, we draw inspiration from a technique proposed in \citep{balkanski2016power}, estimating the value of $\mathbb{E}_{\Pi_{t+1, i}}\big[F(\Pi_{t+1, i})\big]$ and  $\mathbb{E}_{\Pi_{t,-i}}\big[F(\Pi_{t,-i})\big]$ using $\verb"avg"(\Phi_{t+1, i})$ and $\verb"avg"(\Phi_{t,-i})$ respectively. Here, $\verb"avg"(\Phi_{t+1, i})$ represents the average (observed) utility of all samples where the length is $t+1$ and $i$ is placed at the last slot, while $\verb"avg"(\Phi_{t,-i})$ denotes the average (observed) utility of all samples with length $t$ that do not contain $i$. Then we use
\begin{eqnarray}\widetilde{\Delta}(i,t) = \verb"avg"(\Phi_{t+1, i}) - \verb"avg"(\Phi_{t,-i})\end{eqnarray}
as an estimation of $\Delta(i, t)$ for all $i\in \Omega$ and $t\in [k-1]$.

In the following, we treat $\widetilde{\Delta}(i,t)$ as the weight of placing $i$ at position $t+1$. As a subroutine of our algorithm, we aim to find a feasible sequence that maximizes the total weight. This objective can be reframed as a \emph{maximum weight matching problem}. Specifically, we introduce a set of item-position pairs $\Psi=\{(i, t)\mid i\in \Omega, t\in \{1, 2, \cdots, k\}\}$, where selecting a pair $(i, t)$ indicates assigning item $i$ to position $t$. Consequently, the task of identifying a feasible sequence maximizing the total weight is transformed into the following maximum weight matching problem.
 \begin{center}
\framebox[0.4\textwidth][c]{
\enspace
\begin{minipage}[t]{0.4\textwidth}
\small
$\textbf{P.1}$
$\max_{\psi \subseteq\Psi: |\psi|\leq k} \sum_{(i,t)\in \psi} \widetilde{\Delta}(i,t-1)$ \\
\textbf{subject to} $|\psi \cap \Psi_i|\leq 1$  for all $i\in \Omega;$ \\
$|\psi \cap \Psi_t|= 1$ for all $t\in [k-1]$.
\end{minipage}
}
\end{center}
\vspace{0.1in}

Here $\Psi_i=\{(i,t) \mid t\in \{1, 2, \cdots, k\}\}$ denote the set of all item-position pairs involving item $i$, and $\Psi_t=\{(i,t) \mid i\in \Omega\}$ denote the set of all item-position pairs involving position $t$. The condition ``$|\psi \cap \Psi_i|\leq 1$ for all $i\in \Omega$'' ensures that each item appears at most once in a sequence, while ``$|\psi \cap \Psi_t|= 1$ for all $t\in [k-1]$'' ensures that each position contains exactly one item. It is straightforward to confirm the existence of a one-to-one correspondence between feasible sequences and feasible solutions of $\textbf{P.1}$. That is, given a feasible solution $\psi$ of $\textbf{P.1}$, one can construct a feasible sequence such that for each $i\in \Omega$ and $t\in \{1, 2, \cdots, k\}$, item $i$ is placed in position $t$ if and only if $(i,t)\in \psi$.

Because $\textbf{P.1}$ is a classic maximum weighted matching problem, it can be solved efficiently in polynomial time \cite{schrijver2003combinatorial}. Now we are ready to present our final algorithm  (as listed in Algorithm \ref{alg:2}). First, we solve $\textbf{P.1}$ optimally, and let $\pi^s$ denote the sequence corresponding to this solution. Then, we compute the final sequence as follows:  If $(1-c)^2 \geq \alpha\cdot \frac{1-c}{1+c-c^2}$, where $\alpha=\frac{n-k}{n}\cdot \frac{n-k-1}{n-1} \cdot \ldots \cdot  \frac{n-2k+1}{n-k+1}$, then our algorithm returns $\pi^s$ as the final solution. Otherwise, if  $(1-c)^2 < \alpha\cdot \frac{1-c}{1+c-c^2}$ and $(1-c)\sum_{t\in\{1, \cdots, k\}} \widetilde{\Delta}(\pi^s_t,t-1)\geq  \verb"avg"(\Phi_k)$, then our algorithm still returns $\pi^s$ as the final solution. Here, $\text{avg}(\Phi_k)$ denotes the average utility of all samples with a sequence length of $k$. Otherwise, our algorithm returns a random sequence of length $k$ as the final solution.

Note that our algorithm requires the curvature \( c \) of each individual function as input. If \( c \) is unknown, we can adopt \( \pi^s \) as our final solution, yielding an approximation ratio of \( (1 - c)^2 \) (please refer to \textbf{Discussions} section for more details).

\begin{algorithm}[hptb]
\caption{Sequencing-from-Samples}
\label{alg:2}
\begin{algorithmic}[1]
\STATE Solve $\textbf{P.1}$ to obtain $\pi^s$
\IF{$(1-c)^2 \geq \alpha\cdot \frac{1-c}{1+c-c^2}$} \label{lm:1}
    \STATE $\pi^\diamond \leftarrow \pi^s$
\ELSIF{$(1-c)\sum_{t\in\{1, \cdots, k\}} \widetilde{\Delta}(\pi^s_t,t-1)\geq \text{avg}(\Phi_k)$} \label{lm:2}
    \STATE $\pi^\diamond \leftarrow \pi^s$
\ELSE
    \STATE $\pi^\diamond \leftarrow$ a random sequence of length $k$
\ENDIF
\RETURN $\pi^\diamond$; \label{l:1}
\end{algorithmic}
\end{algorithm}

\section{Performance Analysis}
Let $\pi^\diamond$ be the sequence returned from Algorithm \ref{alg:2}, we next analyze the approximation ratio of $\pi^\diamond$, assuming $f_t$ is a monotone submodular function with curvature $c$ for all $t\in \{1, 2, \cdots, k\}$. We first present two technical lemmas. The first lemma derives an approximation ratio for the case when $(1 - c)^2 \geq \alpha \cdot \frac{1 - c}{1 + c - c^2}$, while the second lemma derives an approximation ratio for the remaining cases. The final approximation ratio is the better of these values.
\begin{lemma}
\label{lem:a}
Assume $f_t$ is a monotone submodular function with curvature $c$ for all $t\in \{1, 2, \cdots, k\}$, for the case when $(1-c)^2 \geq \alpha\cdot \frac{1-c}{1+c-c^2}$, we have that,  with a sufficiently large
polynomial number of samples,
\begin{eqnarray}
F(\pi^\diamond) \geq  \big((1-c)^2-o(1)\big)F(\pi^*)
\end{eqnarray} where $\alpha=\frac{n-k}{n}\cdot \frac{n-k-1}{n-1} \cdot \ldots \cdot  \frac{n-2k+1}{n-k+1}$.
\end{lemma}
\emph{Proof:} According to Line \ref{lm:1} in Algorithm \ref{alg:2}, when  $(1-c)^2 \geq \alpha\cdot \frac{1-c}{1+c-c^2}$, it returns $\pi^s$ as $\pi^\diamond$. Here $\pi^s$ denotes the sequence corresponding to the optimal solution of $\textbf{P.1}$. To prove this lemma, it suffices to show that $F(\pi^s) \geq  \big((1-c)^2-o(1)\big)F(\pi^*)$.

Let $\pi^s=\{e_1, e_2, \cdots, e_k\}$ and $\pi^s_{[t]}=\{e_1, e_2, \cdots, e_t\}$,  it follows that
{\small\begin{eqnarray*}
&&F(\pi^s) = \sum_{t\in[k-1]} F(\pi^s_{[t+1]}) - F(\pi^s_{[t]})\\
&&= \sum_{t\in[k-1]} \sum_{j\in\{t+1, \cdots, k\}}  f_j(e_{t+1}\mid \pi^s_{[t]})\\
&&\geq (1-c) \sum_{t\in[k-1]} \sum_{j\in\{t+1, \cdots, k\}}  f_j(e_{t+1})\\
&&\geq  (1-c) \sum_{t\in[k-1]} \sum_{j\in\{t+1, \cdots, k\}}  \mathbb{E}_{R_{t,-e_{t+1}}}\big[f_j(e_{t+1}\mid R_{t,-e_{t+1}})\big]\\
&&=  (1-c) \sum_{t\in[k-1]} \bigg(\mathbb{E}_{\Pi_{t+1, e_{t+1}}}\big[F(\Pi_{t+1, e_{t+1}})\big] \\
&&\quad\quad\quad\quad\quad\quad\quad\quad- \mathbb{E}_{\Pi_{t,-e_{t+1}}}\big[F(\Pi_{t,-e_{t+1}})\big]\bigg)\\
&&= (1-c) \sum_{t\in[k-1]} \Delta(e_{t+1},t)
\end{eqnarray*}}
where $R_{t,-e_{t+1}}$ denotes a random set of size $t$ that excludes item $e_{t+1}$. The first inequality is by the curvature of $f_t$ and fact that $e_{t+1}\notin  \pi^s_{[t]}$ for all $t\in [t-1]$, and the second inequality is by the assumption that $f_t$ is submodular for all $t\in \{1, 2, \cdots, k\}$.

Recall that $\Delta(i, t)= \mathbb{E}_{\Pi_{t+1, i}}\big[F(\Pi_{t+1, i})\big] - \mathbb{E}_{\Pi_{t,-i}}\big[F(\Pi_{t,-i})\big]
$ and $\widetilde{\Delta}(i,t) = \verb"avg"(\Phi_{t+1, i}) - \verb"avg"(\Phi_{t,-i})$ is an estimation of $\Delta(i, t)$. In the appendix (Lemma \ref{lem:appenx}), we show that with a sufficiently large polynomial number of samples, the estimation $\widetilde{\Delta}(i,t)$ is $n^2$-close to $\Delta(i,t)$ for all $i\in \Omega$ and $t\in[k-1]$, with high probability, i.e.,
\begin{eqnarray}
\label{eq:7}
\Delta(i,t)+\frac{\delta}{n^2} \geq \widetilde{\Delta}(i,t) \geq \Delta(i,t)-\frac{\delta}{n^2}.
\end{eqnarray}
where $\delta=\max_{\pi: |\pi|\leq k} \phi(\pi)$ denotes the maximum  realized value of any sequence with a length of at most $k$. Recall that in the example of product sequencing, $\phi(\pi)=1$ indicates a purchase, while $\phi(\pi)=0$ indicates a non-purchase. Therefore, in this example, $\delta=1$.

This, together with the previous inequality, implies that
\begin{eqnarray}
&&F(\pi^s)\geq (1-c) \sum_{t\in[k-1]} \Delta(e_{t+1},t) ~\nonumber\\
&&\geq  (1-c) \sum_{t\in[k-1]} \widetilde{\Delta}(e_{t+1},t) - \frac{\delta}{n}.\label{eq:3}
\end{eqnarray}

Recall that $\pi^s=\{e_1, e_2, \cdots, e_k\}$ is the sequence corresponding to the optimal solution of $\textbf{P.1}$, we have
\begin{eqnarray}
\label{eq:14}
 &&\sum_{t\in[k-1]} \widetilde{\Delta}(e_{t+1},t)\geq  \sum_{t\in[k-1]} \widetilde{\Delta}(e^*_{t+1},t)~\nonumber\\
  &&\geq  \sum_{t\in[k-1]} \Delta(e^*_{t+1},t)-\frac{\delta}{n}.
\end{eqnarray}
Here the second inequality is derived using inequality (\ref{eq:7}).

In addition, observe that
\begin{eqnarray*}
&&\sum_{t\in[k-1]} \Delta(e^*_{t+1},t) =\sum_{t\in[k-1]} \bigg(\mathbb{E}_{\Pi_{t+1, e^*_{t+1}}}\big[F(\Pi_{t+1, e^*_{t+1}})\big]\\
 &&\quad\quad\quad\quad\quad\quad\quad\quad\quad\quad - \mathbb{E}_{\Pi_{t,-e^*_{t+1}}}\big[F(\Pi_{t,-e^*_{t+1}})\big]\bigg)\\
&&= \sum_{t\in[k-1]} \sum_{j\in\{t+1, \cdots, k\}}  \mathbb{E}_{R_{t,-e^*_{t+1}}}\big[f_j(e^*_{t+1}\mid R_{t,-e^*_{t+1}})\big]\\
&&\geq \sum_{t\in[k-1]} \sum_{j\in\{t+1, \cdots, k\}} \mathbb{E}_{R_{t,-e^*_{t+1}}}\big[(1-c)f_j(e^*_{t+1})\big]\\
&&= (1-c) \sum_{t\in[k-1]} \sum_{j\in\{t+1, \cdots, k\}} f_j(e^*_{t+1})\\
&&\geq (1-c) F(\pi^*)
\end{eqnarray*}
where the first inequality is by the curvature of $f_t$ and fact that $e^*_{t+1}\notin  R_{t,-e^*_{t+1}}$  for all $t\in [k-1]$, and the second inequality is by the assumption that  $f_t$ is submodular for all $t\in \{1, 2, \cdots, k\}$ .

This, together with inequality (\ref{eq:14}), implies that
\begin{eqnarray}
 \sum_{t\in[k-1]} \widetilde{\Delta}(e_{t+1},t) &&\geq \sum_{t\in[k-1]} \Delta(e^*_{t+1},t)-\frac{\delta}{n} ~\nonumber\\
 &&\geq (1-c) F(\pi^*)-\frac{\delta}{n}. \label{eq:4}
\end{eqnarray}

Inequalities (\ref{eq:3}) and (\ref{eq:4}) imply that
\begin{eqnarray}
\label{eq:5}
F(\pi^s) \geq  \big((1-c)^2-o(1)\big)F(\pi^*).
\end{eqnarray} $\Box$


We proceed to providing the second technical lemma.

\begin{lemma}
\label{lem:b}
Assume $f_t$ is a monotone submodular function with curvature $c$ for all $t\in \{1, 2, \cdots, k\}$, for the case when $(1-c)^2 < \alpha\cdot \frac{1-c}{1+c-c^2}$, we have that,  with a sufficiently large
polynomial number of samples,
\begin{eqnarray}
F(\pi^\diamond) \geq  \alpha\cdot (\frac{1-c}{1+c-c^2}- o(1)) F(\pi^*)
\end{eqnarray} where $\alpha=\frac{n-k}{n}\cdot \frac{n-k-1}{n-1} \cdot \ldots \cdot  \frac{n-2k+1}{n-k+1}$.
\end{lemma}
\emph{Proof:} Let us define a function $F(\pi \uplus \pi^*)$ for a  sequence $\pi$ of length $k$ and an optimal solution $\pi^*$ as follows:
\begin{eqnarray} F(\pi \uplus \pi^*) = \sum_{t\in[k]} f_t(\pi_{[t]} \cup \pi^*_{[t]}) \end{eqnarray}
Here, $\pi_{[t]}$ (and $\pi^*_{[t]}$) represent all items from $\pi$ (and $\pi^*$) respectively, that are placed up to position $t$. That is, $\pi \uplus \pi^*$ can be envisioned as a virtual sequence where both $\pi_t$ and $\pi^*_t$ are placed at position $t$ for all $t\in \{1, 2, \cdots, k\}$.

Let $\Pi'$ denote a random sequence of length $k$ that is sampled over items from $\Omega\setminus \pi^*$, and $\Pi'_{[t]}$ denotes the first $t$ items from $\Pi'$, observe that,
{\small\begin{eqnarray*}&& \mathbb{E}_{\Pi'}\big[F(\Pi' \uplus \pi^*) -F(\pi^*)\big] \\
&&= \mathbb{E}_{\Pi'}\big[\sum_{t\in \{1, 2, \cdots, k\}}  f_t(\Pi'_{[t]} \cup \pi^*_{[t]}) - \sum_{t\in \{1, 2, \cdots, k\}}  f_t(\pi^*_{[t]})\big] \\
&& =  \mathbb{E}_{\Pi'}\big[\sum_{t\in \{1, 2, \cdots, k\}} ( f_t(\Pi'_{[t]} \cup \pi^*_{[t]}) - f_t(\pi^*_{[t]}))\big]\\
&&=  \mathbb{E}_{\Pi'}\big[\sum_{t\in \{1, 2, \cdots, k\}}   f_t(\Pi'_{[t]} \mid \pi^*_{[t]})\big]\\
&&= \sum_{t\in \{1, 2, \cdots, k\}}   \mathbb{E}_{\Pi'}\big[ f_t(\Pi'_{[t]} \mid \pi^*_{[t]})\big]\\
&&\geq \sum_{t\in \{1, 2, \cdots, k\}} (1-c)\mathbb{E}_{\Pi'}[f_t(\Pi'_{[t]})] \\
&& \geq (1-c)\mathbb{E}_{\Pi'}[ \sum_{t\in \{1, 2, \cdots, k\}}  f_t(\Pi'_{[t]})] = (1-c)\mathbb{E}_{\Pi'}[F(\Pi')]. \end{eqnarray*}}
To establish the first inequality, we utilize the fact that $\Pi'_{[t]}$ is a random set of size $t$ and $\Pi'_{[t]} \subseteq \Omega \setminus \pi^*_{[t]}$. Consequently, this inequality can be derived by substituting $R = \Pi'_{[t]}$ and $S = \pi^*_{[t]}$ into Lemma \ref{lem:11} which is presented in the appendix.


In addition, observe that
\begin{eqnarray*}
&&F(\pi^*) + \mathbb{E}_{\Pi'}[F(\Pi' \uplus \pi^*)] -F(\pi^*) \\
&&= \mathbb{E}_{\Pi'}[F(\Pi')] + \mathbb{E}_{\Pi'}[F(\Pi' \uplus \pi^*)] -\mathbb{E}_{\Pi'}[F(\Pi')] \end{eqnarray*}
and $\sum_{t\in[k-1]} \Delta(e^*_{t+1},t)\geq \alpha\cdot \mathbb{E}_{\Pi'}\big[F(\Pi' \uplus \pi^*) -F(\Pi')\big]$ where $\alpha=\frac{n-k}{n}\cdot \frac{n-k-1}{n-1} \cdot \ldots \cdot  \frac{n-2k+1}{n-k+1}$ (by Lemma \ref{lem:1} in the appendix). We have
\begin{eqnarray*}
&&F(\pi^*) + \mathbb{E}_{\Pi'}[F(\Pi' \uplus \pi^*)] -F(\pi^*)\\
&&\leq \mathbb{E}_{\Pi'}[F(\Pi')] + \frac{1}{\alpha}\sum_{t\in[k-1]} \Delta(e^*_{t+1},t).\end{eqnarray*}

This, together with the previous observation that $\mathbb{E}_{\Pi'}[F(\Pi' \uplus \pi^*) -F(\pi^*)]\geq(1-c)\mathbb{E}_{\Pi'}[F(\Pi')]$, implies that
$
F(\pi^*) + (1-c)\mathbb{E}_{\Pi'}[F(\Pi')] \leq \mathbb{E}_{\Pi'}[F(\Pi')] + \frac{1}{\alpha} \sum_{t\in[k-1]} \Delta(e^*_{t+1},t)$.
It follows that
\begin{eqnarray}
\sum_{t\in[k-1]} \Delta(e^*_{t+1},t) \geq \alpha \bigg(1-c\frac{ \mathbb{E}_{\Pi'}[F(\Pi')]}{F(\pi^*)}\bigg)F(\pi^*).\label{eq:2}\end{eqnarray}

This, together with inequality (\ref{eq:4}), implies that
{\small\begin{eqnarray}
&&(1-c)\sum_{t\in[k-1]} \widetilde{\Delta}(e_{t+1},t)~\nonumber \\
&&\geq  (1-o(1))(1-c)\sum_{t\in[k-1]} \Delta(e^*_{t+1},t) ~\nonumber\\
&&\geq (1- o(1))(1-c)\alpha \bigg(1-c\frac{ \mathbb{E}_{\Pi'}[F(\Pi')]}{F(\pi^*)}\bigg)F(\pi^*).\label{eq:6}\end{eqnarray}}

According to Line \ref{lm:2} of Algorithm \ref{alg:2} and inequality (\ref{eq:3}), when $(1-c)^2 < \alpha\cdot \frac{1-c}{1+c-c^2}$, $\pi^\diamond$ achieves an utility of at least $ \max\{(1-o(1))\mathbb{E}_{\Pi}[F(\Pi)],  (1-o(1))(1-c)\sum_{t\in[k-1]} \widetilde{\Delta}(e_{t+1},t)\}$ where $\Pi$ denotes a random sequence of length $k$ that is sampled over items from $\Omega$. Hence, the approximation ratio of our algorithm is at least  $\max\{(1-o(1)) \frac{\mathbb{E}_{\Pi}[F(\Pi)]}{F(\pi^*)}, (1-o(1)) \frac{(1-c)\sum_{t\in[k-1]} \widetilde{\Delta}(e_{t+1},t)}{F(\pi^*)}\}$. According to inequality (\ref{eq:6}), $\frac{(1-c)\sum_{t\in[k-1]} \widetilde{\Delta}(e_{t+1},t)}{F(\pi^*)}\geq (1-o(1))\alpha (1-c)(1-c\frac{ \mathbb{E}_{\Pi'}[F(\Pi')]}{F(\pi^*)})$. It follows that the approximation ratio of our algorithm is at least $\max\{(1-o(1))\frac{\mathbb{E}_{\Pi}[F(\Pi)]}{F(\pi^*)}, (1-o(1))\alpha(1-c)(1-c\frac{ \mathbb{E}_{\Pi'}[F(\Pi')]}{F(\pi^*)})\}=(1-o(1))\max\{\frac{\mathbb{E}_{\Pi}[F(\Pi)]}{F(\pi^*)}, \alpha(1-c)(1-c\frac{ \mathbb{E}_{\Pi'}[F(\Pi')]}{F(\pi^*)})\}\geq (1-o(1))\max\{\frac{\alpha\mathbb{E}_{\Pi'}[F(\Pi')]}{F(\pi^*)}, \alpha(1-c)(1-c\frac{ \mathbb{E}_{\Pi'}[F(\Pi')]}{F(\pi^*)})\}=(1-o(1))\alpha\max\{\frac{\mathbb{E}_{\Pi'}[F(\Pi')]}{F(\pi^*)}, (1-c)(1-c\frac{ \mathbb{E}_{\Pi'}[F(\Pi')]}{F(\pi^*)})\}$ where the inequality is by the observation that  the probability that $\Pi$ is sampled from $\Omega\setminus \pi^*$ is at least $\alpha=\frac{n-k}{n}\cdot \frac{n-k-1}{n-1} \cdot \ldots \cdot  \frac{n-2k+1}{n-k+1}$. Observe that $\max\{\frac{\mathbb{E}_{\Pi'}[F(\Pi')]}{F(\pi^*)}, (1-c)(1-c\frac{ \mathbb{E}_{\Pi'}[F(\Pi')]}{F(\pi^*)})\}$ is at least $\frac{1-c}{1+c-c^2}$, hence, the approximation of $\pi^\diamond$ is at least $\alpha\cdot \frac{1-c}{1+c-c^2}- o(1)$. $\Box$

Combining Lemma \ref{lem:a} and Lemma \ref{lem:b}, we have the following theorem.

\begin{theorem} Let $\pi^\diamond$ be the sequence returned from Algorithm \ref{alg:2}, assuming $f_t$ is a monotone submodular function with curvature $c$ for all $t\in \{1, 2, \cdots, k\}$, we have that,  with a sufficiently large
polynomial number of samples,
{\small\begin{eqnarray}
F(\pi^\diamond) \geq \max\{(1-c)^2-o(1), \alpha\cdot \frac{1-c}{1+c-c^2}- o(1)\} F(\pi^*)
\end{eqnarray}} where $\alpha=\frac{n-k}{n}\cdot \frac{n-k-1}{n-1} \cdot \ldots \cdot  \frac{n-2k+1}{n-k+1}$.
\end{theorem}

\subsection{Discussions}
\label{sec:discussiopn}
We present two remarks: one regarding the design of our algorithm and the other addressing a potential gap in existing studies.

First, our algorithm design and analysis assume a good estimation of the curvature $c$ of each individual function. This assumption might not always hold; if $c$ is unknown, we can adopt $\pi^s$ as our final solution, yielding an approximation ratio of $(1-c)^2$, as shown in Lemma \ref{lem:a}.

Second, while our study builds on the work of \cite{balkanski2016power} by extending the ``learning-from-samples'' approach from set functions to sequence functions, we identify a potential gap in their  analysis. Specifically, their proof of Lemma 1 relies on the assumption that
$f(R \mid  S^\star ) \geq (1-c)f(R)$,
where $S^\star$ is an optimal set solution, $R$ is a uniformly random set of size $k-1$ (with $k$ being the size constraint of the final solution) and $c$ is the curvature of function $f$. This assumption is, unfortunately, not generally valid; according to the definition of the curvature $c$, this assumption holds only if $R \cap S^\star = \emptyset$. Our study addresses this issue by introducing the notion of $\alpha$ and further extends their research to a more complex sequence function.

\section{Appendix}

\begin{lemma}
\label{lem:appenx}
With a sufficiently large polynomial number of samples, the estimation $\widetilde{\Delta}(i,t)$ is $n^2$-close to $\Delta(i,t)$ for all $i\in \Omega$ and $t\in[k-1]$, with high probability, i.e.,
$
\Delta(i,t)+\frac{\delta}{n^2} \geq \widetilde{\Delta}(i,t) \geq \Delta(i,t)-\frac{\delta}{n^2}$ where $\delta=\max_{\pi: |\pi|\leq k} \phi(\pi)$ denotes the maximum  realized value of any sequence with a length of at most $k$.
\end{lemma}
\emph{Proof:} Our proof is inspired by the one presented in \cite{balkanski2016power} (Appendix A); however, we extend their analysis from set functions to sequence functions. Consider an arbitrary pair of $i\in \Omega$ and $t\in[k-1]$.

Observation 1: The probability of sampling a sequence of length $t$ is no less than $1/k$, whose value is at least $1/n$. Note that the case when $t = 0$ is trivial because the value of an empty sequence is known to be zero.  Furthermore, given that the sampled sequence has a length of $t$, the probability of it not containing item $i$ is at least $1-t/n \geq 1/n$. Hence, the probability of sampling a sequence of length $t$ without $i$ is at least $1/n^2$.

Observation 2: The probability of sampling a sequence of length $t+1$ is no less than $1/k$, where $1/k$ is at least $1/n$. Additionally, given that the sampled sequence has a length of $t+1$, the likelihood of the last item being $i$ is at least $1/n$. Consequently, the probability of sampling a sequence of length $t+1$ with $i$ at position $t+1$ is at least $1/n^2$.

The above two observations, together with Chernoff bounds, imply that gathering a minimum of $n^5$ samples of length $t$ that do not contain $i$, and at least $n^5$ samples of length $t+1$ wherein $i$ resides at position $t+1$, can be accomplished with high probability by obtaining $n^8$ samples.

By Hoeffding's inequality and the fact that $\delta$ is the largest possible value observed from any sequence of size at most $k$, we have
$\Pr[|\verb"avg"(\pi_{t,-i}) - \mathbb{E}_{\Pi_{t,-i}}[F(\Pi_{t,-i})]|\geq \frac{\delta}{2n^2}]\leq 2e^{-2n^5(\delta/2n^2)^2/\delta^2} \leq 2e^{-n/2}$,
and
$
\Pr[|\verb"avg"(\pi_{t+1, i}) - \mathbb{E}_{\Pi_{t+1, i}}[F(\Pi_{t+1, i})]|\geq \frac{\delta}{2n^2}] \leq  2e^{-n/2}$.

Given that $\Delta(i,t) = \mathbb{E}_{\Pi_{t+1, i}}[F(\Pi_{t+1, i})] - \mathbb{E}_{\Pi_{t,-i}}[F(\Pi_{t,-i})]$ and $\widetilde{\Delta}(i,t) = \verb"avg"(\pi_{t+1, i}) - \verb"avg"(\pi_{t,-i})$, we can deduce that, with a sample size of $n^8$, the following inequalities hold for all $i \in \Omega$ and $t \in [k-1]$, with high probability:
$\Delta(i,t) + \frac{\delta}{n^2} \geq \widetilde{\Delta}(i,t) \geq \Delta(i,t) - \frac{\delta}{n^2}$. $\Box$

\begin{lemma}
\label{lem:11}
Let $f: 2^\Omega\rightarrow \mathbb{R}_{\geq0}$ be a monotone and submodular function, given  any subset of items $S\subseteq\Omega$ such that $|S|\leq k$,  let $R$ be a set of size $t$ that is  randomly sampled from $\Omega\setminus S$, for any $t\leq \min\{k, |\Omega\setminus S|\}$,
$
\mathbb{E}_R[f(R \mid S)]\geq (1-c)\mathbb{E}_R[f(R)]$.
\end{lemma}

\emph{Proof:} Assuming $R$ is obtained by sequentially sampling $t$ items without replacement, let $R=\{r_1, \cdots, r_t\}$, where $r_j$ represents the $j$-th sampled item. Let $R_{[j]}=\{r_1, \cdots, r_j\}$ denote the first $j$ sampled items,
\begin{eqnarray}
\mathbb{E}_R[f(R \mid S)]= \sum_{j\in[t-1]} \mathbb{E}_R[f(r_{j+1} \mid R_{[j]}\cup S )].
\end{eqnarray}

Consider any given sample $R$, because $r_{j+1}\notin  R_{[j]}$ and $r_{j+1}\notin  S$ (by the assumption that $R\subseteq \Omega\setminus S$), then by the curvature of $f$,  $ f(r_{j+1} \mid R_{[j]}\cup S )\geq (1-c) f(r_{j+1})$. 
It follows that
$\mathbb{E}_R[f(R \mid S)]=  \mathbb{E}_R[\sum_{j\in[t-1]} f(r_{j+1} \mid R_{[j]}\cup S )]
= \sum_{j\in[t-1]} \mathbb{E}_R[f(r_{j+1} \mid R_{[j]}\cup S )]\geq \sum_{j\in[t-1]} (1-c) \mathbb{E}_R[f(r_{j+1})]
 = (1-c)  \mathbb{E}_R[\sum_{j\in[t-1]}  f(r_{j+1})]\geq (1-c)  \mathbb{E}_R[f(R)]$ where the first inequality is by the observation that $ f(r_{j+1} \mid R_{[j]}\cup S )\geq (1-c) f(r_{j+1})$ for any $R$ and  the last inequality is by the assumption that $f$ is a submodular function. $\Box$

\begin{lemma}\label{lem:1} Let $\Pi'$ denote a random sequence of length $k$ that is sampled over items from $\Omega\setminus \pi^*$ where $\pi^*=\{e^*_1, \cdots, e^*_k\}$ denotes the optimal solution, we have
$\sum_{t\in[k-1]} \Delta(e^*_{t+1},t)\geq \alpha\cdot \mathbb{E}_{\Pi'}[F(\Pi' \uplus \pi^*) -F(\Pi')]$ where $\alpha=\frac{n-k}{n}\cdot \frac{n-k-1}{n-1} \cdot \ldots \cdot  \frac{n-2k+1}{n-k+1}$.
\end{lemma}
\emph{Proof:} Let $\Pi$ denote a random sequence of length $k$ that is sampled over items from $\Omega$. Hence, the probability that the first $t$ items $\Pi_{[t]}$ is sampled from $\Omega\setminus \pi^*$ is at least $\alpha=\frac{n-k}{n}\cdot \frac{n-k-1}{n-1} \cdot \ldots \cdot  \frac{n-2k+1}{n-k+1}$ for any $t\in \{1, \cdots, k\}$. Recall that $\Pi'$ denotes a random sequence of length $k$ that is sampled over items from $\Omega\setminus \pi^*$. It follows that $\mathbb{E}_{\Pi}[f_t(i \mid \Pi_{[t]})] \geq \alpha \mathbb{E}_{\Pi'}[f_t(i\mid \Pi'_{[t]})]$ for all $t\in \{1, \cdots, k\}$ and any item $i\in \Omega$.

Observe that $\sum_{t\in[k-1]} \Delta(e^*_{t+1},t)$
\begin{eqnarray*}
&&= \sum_{t\in[k-1]} \mathbb{E}_{\Pi_{t+1, e^*_{t+1}}}[F(\Pi_{t+1, e^*_{t+1}})] - \mathbb{E}_{\Pi_{t,-e^*_{t+1}}}[F(\Pi_{t,-e^*_{t+1}})] \\
&&= \sum_{t\in[k-1]} \mathbb{E}_{\Pi_{t,-e^*_{t+1}}}[\sum_{z\in\{t+1, \cdots, k\}}f_z(e^*_{t+1} \mid \Pi_{t,-e^*_{t+1}})] \\
&&\geq \sum_{t\in[k-1]} \mathbb{E}_{\Pi}[\sum_{z\in\{t+1, \cdots, k\}} f_z(e^*_{t+1} \mid \Pi_{[t]})]\\
&&\geq \sum_{t\in[k-1]} \mathbb{E}_{\Pi}[\sum_{z\in\{t+1, \cdots, k\}} f_z(e^*_{t+1} \mid \Pi_{[z]})]\\
&&= \sum_{t\in[k-1]} \sum_{z\in\{t+1, \cdots, k\}}\mathbb{E}_{\Pi}[f_z(e^*_{t+1} \mid \Pi_{[z]})]\\
&&\geq \sum_{t\in[k-1]}\sum_{z\in\{t+1, \cdots, k\}}\alpha  \mathbb{E}_{\Pi'}[f_z(e^*_{t+1} \mid \Pi'_{[z]})]\\
&&= \alpha  \mathbb{E}_{\Pi'}[\sum_{t\in[k-1]}\sum_{z\in\{t+1, \cdots, k\}} f_z(e^*_{t+1} \mid \Pi'_{[z]})]\\
&&\geq \alpha \mathbb{E}_{\Pi'}[F(\Pi' \uplus \pi^*) -F(\Pi')]
\end{eqnarray*}
where the forth inequality is by the previous observation that $\mathbb{E}_{\Pi}[f_t(i \mid \Pi_{[t]})] \geq \alpha \mathbb{E}_{\Pi'}[f_t(i\mid \Pi'_{[t]})]$ for all $t\in \{1, \cdots, k\}$. $\Box$
\bibliographystyle{ijocv081}
\bibliography{reference}




\end{document}